\documentclass[sigconf]{acmart}

\AtBeginDocument{%
  \providecommand\BibTeX{{%
    \normalfont B\kern-0.5em{\scshape i\kern-0.25em b}\kern-0.8em\TeX}}}

\setcopyright{acmcopyright}
\copyrightyear{2021} 
\acmYear{2021} 
\setcopyright{acmcopyright}\acmConference[SIGSPATIAL '21]{29th International
Conference on Advances in Geographic Information Systems}{November 2--5,
2021}{Beijing, China}
\acmBooktitle{29th International Conference on Advances in Geographic Information
Systems (SIGSPATIAL '21), November 2--5, 2021, Beijing, China}
\acmPrice{15.00}
\acmDOI{10.1145/3474717.3488237}
\acmISBN{978-1-4503-8664-7/21/11}

\begin{document}

\title{Complementary Fusion of Deep Network and Tree Model for ETA Prediction}
\titlenote{* is the corresponding author}

\author{YuRui Huang}
\affiliation{%
 \institution{Nanjing University of Science and Technology}
 \country{China}
 }
\email{huangyurui@njust.edu.cn}

\author{Jie Zhang}
\affiliation{%
 \institution{Huatai Securities}
 \country{China}
 }
\email{zhangjie986862436@gmail.com}

\author{HengDa Bao}
\affiliation{%
 \institution{Baidu}
 \country{China}
 }
\email{baohengda@baidu.com}

\author{Yang Yang $^*$}
\affiliation{%
 \institution{Nanjing University of Science and Technology}
 \country{China}
 }
\email{yyang@njust.edu.cn}

\author{Jian Yang}
\affiliation{%
 \institution{Nanjing University of Science and Technology}
 \country{China}
 }
\email{csjyang@njust.edu.cn}


\begin{abstract}
Estimated time of arrival (ETA) is a very important factor in the transportation system. It has attracted increasing attentions and has been widely used as a basic service in navigation systems and intelligent transportation systems.
In this paper, we propose a novel solution to the ETA estimation problem, which is an ensemble on tree models and neural networks. We proved the accuracy and robustness of the solution on the A/B list and finally won first place in the SIGSPATIAL 2021 GISCUP competition.
\end{abstract}




\maketitle
\section{Introduction}
\begin{figure}
	\begin{center}
		\begin{minipage}[h]{80mm}
			\centering
			\includegraphics[width=\textwidth]{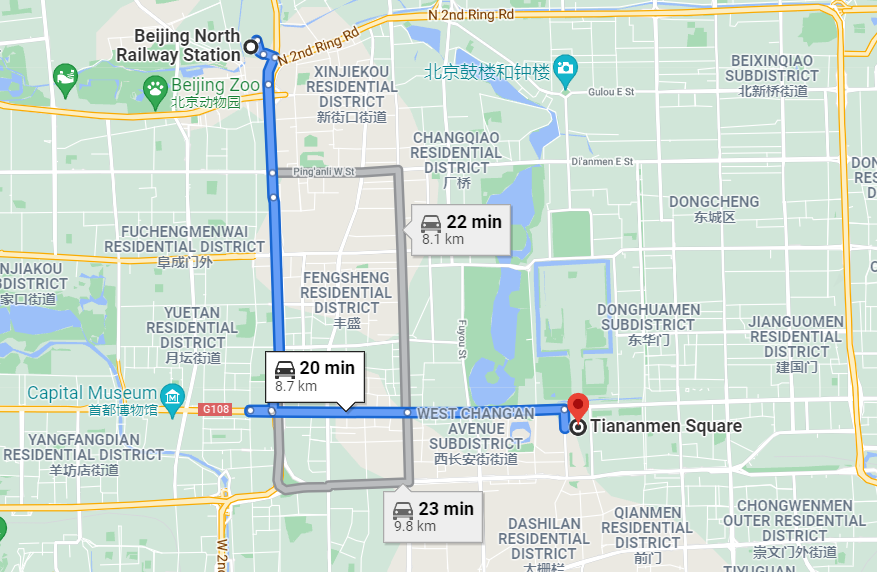}
		\end{minipage}
		\caption{An example of estimated time of arrival (ETA), the white dot represents the starting position, the red point represents the destination position, the blue route represents the shortest travel route in time, and the other two gray lines represent other travel routes.}\label{fig:1}
	\end{center}
\end{figure}

\begin{figure*}[h]
	\centering
	\includegraphics[width=\textwidth]{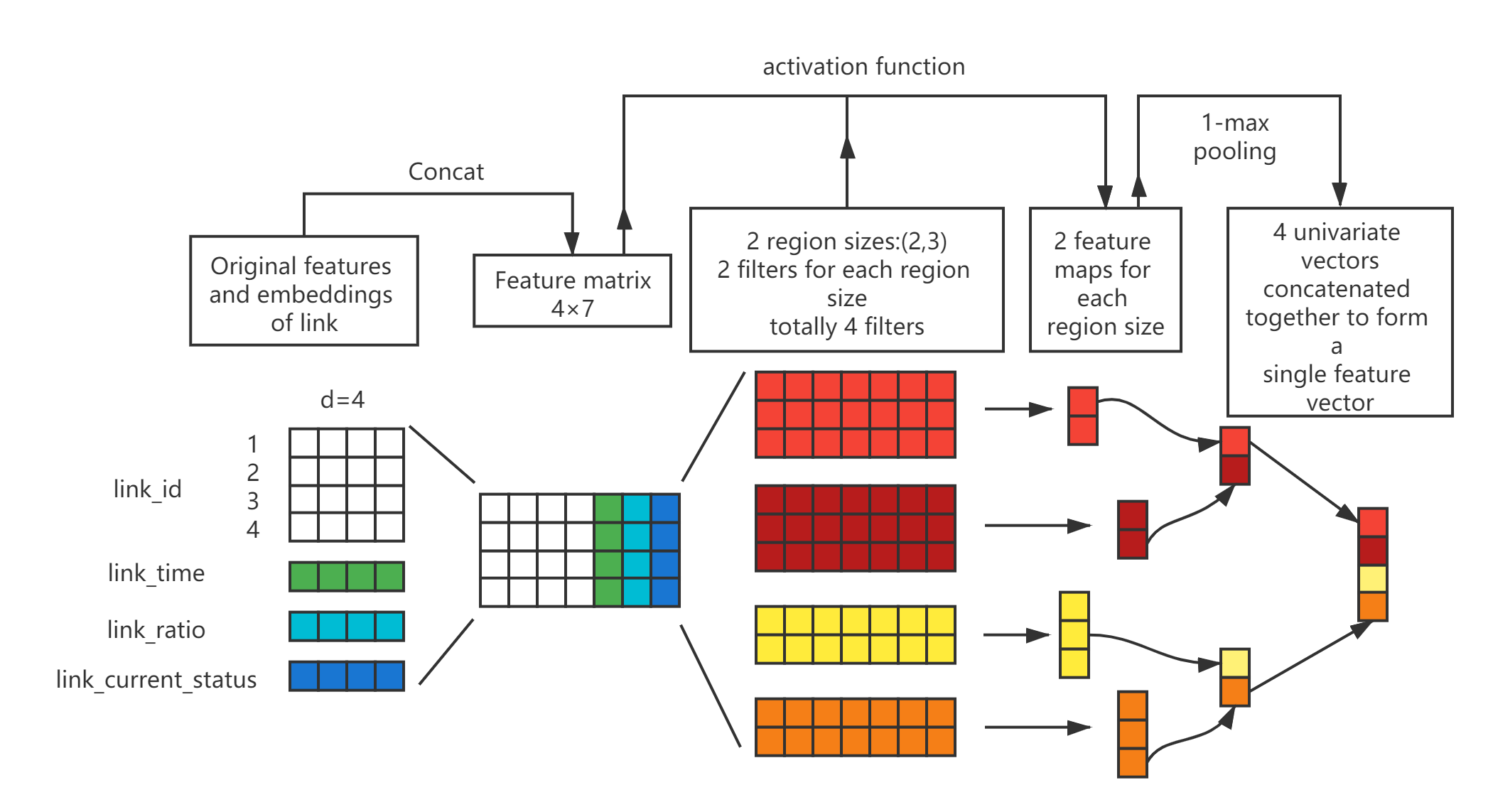}
	\caption{A path sequence containing only 4 links is used to illustrate the convolution structure. Compared with \cite{zhang2015sensitivity}, we add  three-dimensional link features after the embedding layer, and the other processes are the same as TextCNN.}\label{fig:2}
\end{figure*}

Estimated time of arrival (ETA) is a very critical role in the navigation system, which can be considered as a physical quantity of the travel time from the starting point to the destination point. The driving path is composed of a series of road segments. ETA is often used on ride-hailing platforms because travel time is one of the key factors for passengers and drivers to reach a deal. Therefore, ETA is often associated with the user's decision-making, the platform's route planning, the user's itinerary, and the calculation of the fare. Figure 1 shows a real example, which estimates the travel time before the start of the trip. Therefore, an accurate ETA estimate is one of the basic services provided by the online car-hailing platform for users. \cite{yang2017instance}
For ETA estimation problem, we propose a new solution, which is an ensemble based on tree models and neural networks. The neural network uses the CNN structure to capture the information of the upper and lower links, and the A/B list test verifies that the model has a good generalization ability.

\section{APPROACH}
The method of this question is based on the data used in the SIGSPATIAL 2021 GISCUP competition, which is the actual user order itinerary data of Shenzhen in August 2020. It includes data from three aspects: 1)User order route information  2)Road network topology data 3)Weather data. To better improve the performance of the model, we have done rich feature engineering.
\subsection{Feature Engineering}
\textbf{Neural network feature engineering:} The original data has rich information. Features related to the link sequence can be used by the neural network \cite{yang2019semi}. In addition, we have extracted the corresponding dense features and category features.
\begin{enumerate}
\item Dense features: Since ETA is a time prediction problem, the time-related features are particularly important, so we calculated the sum and maximum value of link-time and cross-time \cite{yang2019deep}. In addition, the state of road conditions is a key piece of information. Thus, we have created five quantitative statistics on the state of road conditions. Finally, each piece of data provides us with the distance and the estimated time of the journey, and we calculated an estimated average speed \cite{yang2023towards}.
\item Category characteristics: The driver’s last order time can describe the driver’s order behavior and historical road condition information \cite{yang2019semi2}. Therefore, we have made statistics on the driver’s historical order time, including the characteristics of the driver’s last order time and the driver's second-to-last order time. If the driver does not have the order time, we fill it with -999.
\end{enumerate}

\begin{figure*}[h]
	\centering
	\includegraphics[width=\textwidth]{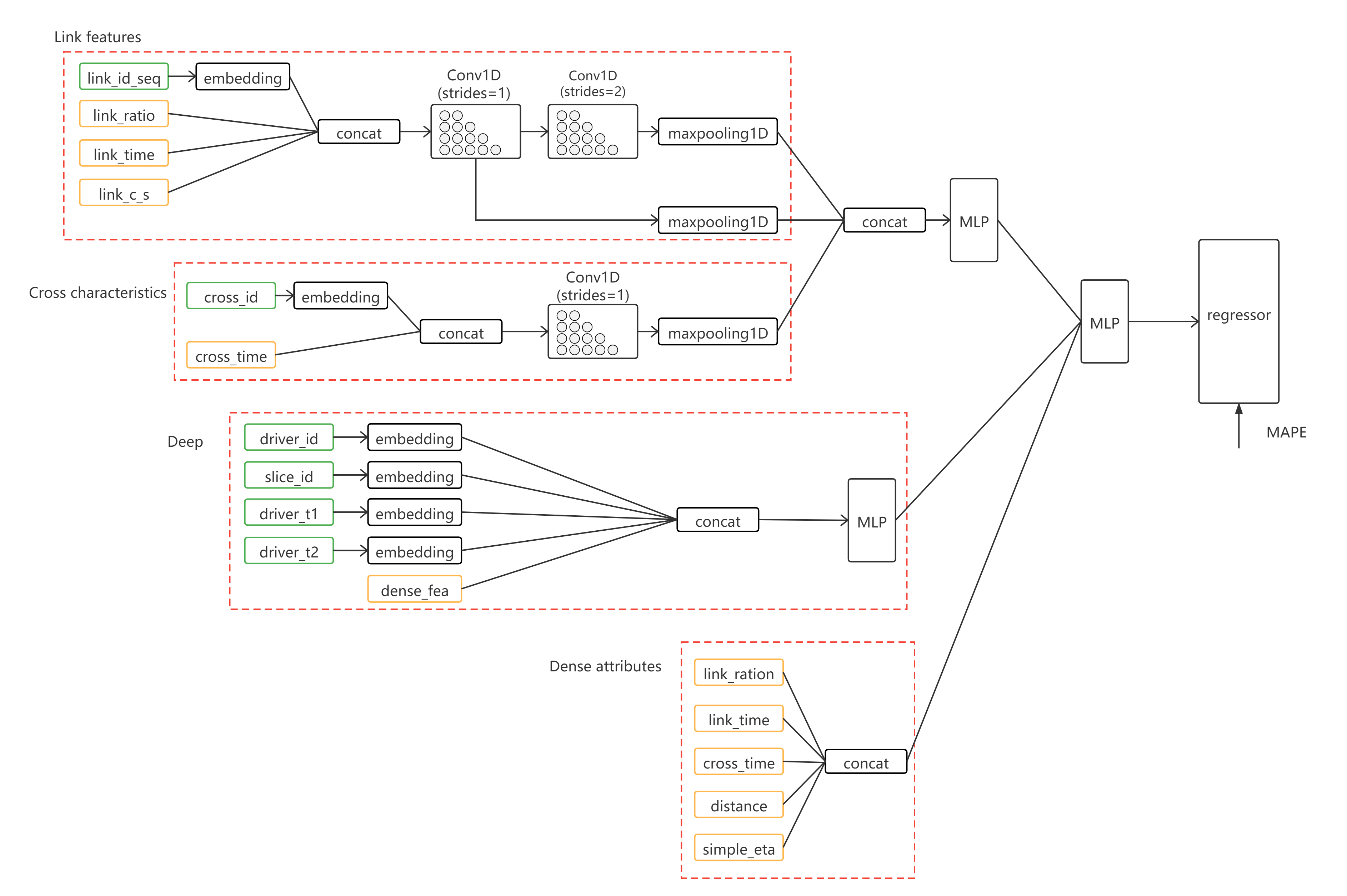}
	\caption{The green box refers to sequence features and category characteristics, the yellow box refers to dense attributes, the convolution kernel uses one-dimensional convolution, and the number of circles represents the length of the convolution kernel.}\label{fig:2}
\end{figure*}

\textbf{Tree model feature engineering:} Since the tree model cannot directly use the original path information (sequence of road segment numbering, estimated travel time sequence of each road segment, road condition sequence, etc.), the method of extracting features is adopted for the characteristics of the tree model \cite{yang2021rethinking}. We summarize the features into several types: statistical features, temporal features, time features, spatial features, and neural network embedding.
\begin{enumerate}
\item For statistical features, we count link-time, cross-time, and the number of links in the same way as neural network features. In addition, we also calculate the average speed, average distance, and congestion distance.
\item As traffic conditions often have tidal phenomena, it is necessary to construct features for time and date. For the date, because the traffic conditions on weekdays and weekends usually differ, we constructed a bool feature of whether it is a weekend or not. For the time, we first constructed the hour feature based on the slice-id (generated every five minutes) and then constructed the feature of whether the time belongs to the rush hours: between 7 am to 9 am or between 17 pm to 19 pm. Finally, the slice-id is divided into early, middle, and late bins.
\item Different road sections have unequal topological structures, so according to the given road network data (nextlinks.txt), construct the upstream and downstream information of each intersection. Then we count the sum of the upstream and downstream quantities of the link in the order.
\item Since the link sequence and the cross sequence have contextual meanings, we use them as text sequences to extract word vector features and LDA features \cite{yuan2021incremental}.
\item In addition, we also add the supervised embedding in the neural network to the features of the tree model. Specifically, we added the following features, such as: link sequence, cross sequence, slice-id, driver-id, the driver's last order time, and the driver's second-to-last order time.
\end{enumerate}

\subsection{Modeling scheme}
\subsubsection{Neural network structure}
The neural network uses the main structure of CNN\cite{lecun1998gradient}. Because the sequence is an one-dimensional structure, we can treat it as "text". For example, in the article \cite{zhang2015sensitivity}, an one-dimensional convolution kernel and global maximum pooling will be used for feature extraction. We denote the dimensionality of the link-id vectors by $d$. If the length of a sequence is $s$, the dimension of the link sequence matrix is $s\times d$. And each link-id has the current description characteristics, including link-time, link-ratio, and link-current-status. We combine it with link Sequence matrix splicing. At this time, the dimension of the sequence matrix becomes $s\times \left ( d + 3 \right )$.

\begin{figure*}[h]
	\centering
	\includegraphics[width=0.75\textwidth]{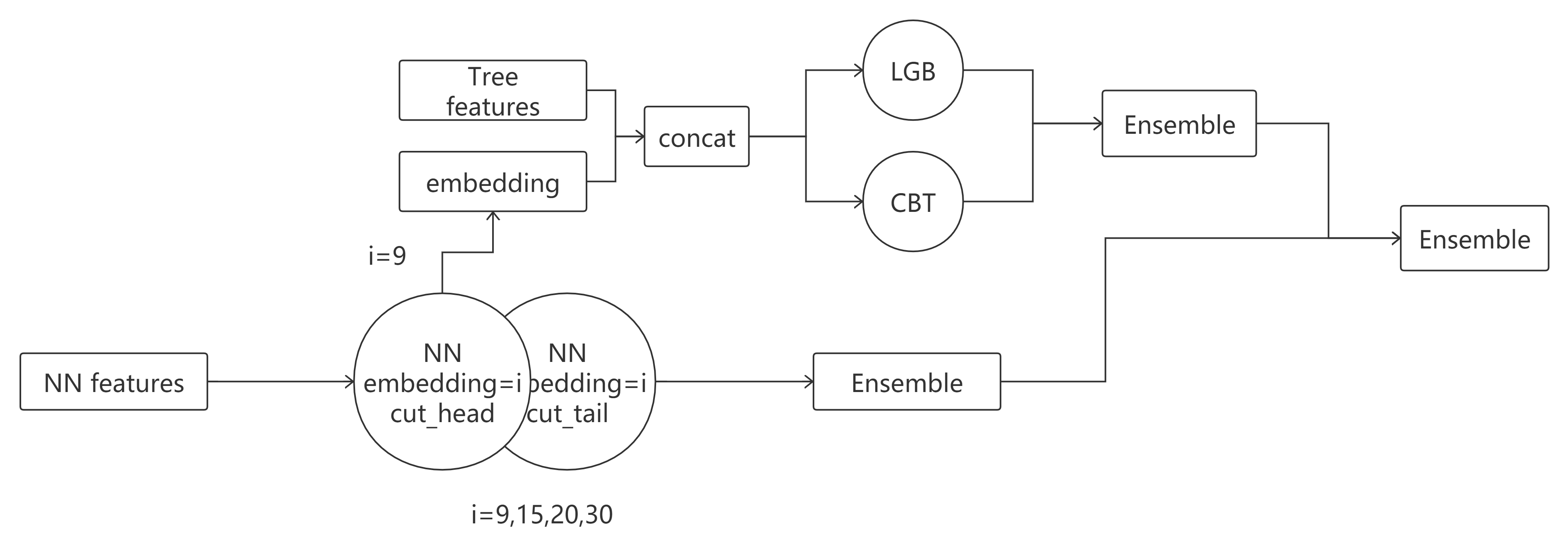}
	\caption{The fusion structure of the tree model and the neural network. The fusion weight can be obtained by the performance of the verification set.}\label{fig:4}
\end{figure*}

Suppose that there is a filer parameterized by the weight matrix $w$ with region size $h$; $w$ will contain $h\cdot d$ parameters to be estimated. We denote the sentence matrix by $M\in \mathbb{R}^{s\times d}$ and use $M\left [ i:j \right ]$ to represent the sub-matrix of $M$ from row $i$ to row $j$. The output sequence $\mathbf{o}\in \mathbb{R}^{s-h+1}$ of the convolution operator is obtained by repeatedly applying the filter on sub-matrices of $M$:
\begin{equation}
    o_{i}=w\cdot M\left [ i:i+h-1 \right ]
\end{equation}
where $i= 1...s-h+1$, and $\cdot $ is the dot product between the sub-matrix and the filter (a sum over element-wise multiplications). We add a bias term $b\in\mathbb{R}$ and an activation function $f$ to each $o_{i}$, inducing the feature map $\mathbf{c}\in \mathbb{R}^{s-h+1}$ for this filter: 
\begin{equation}
    c_{i}=f\left ( o_{i}+b \right )
\end{equation}
For $\mathbf{c}$, we use a one-dimensional global maximum pooling layer for pooling. The process is shown in Figure 2. Specifically, in the convolutional layer, we use one-dimensional convolution kernels of sizes 2, 3, 4, and 5. Each size uses 32 filters, so after pooling, 128-dimensional features will be obtained \cite{zhu2020sequential}. Since the link sequence is relatively long, we set up a two-layer convolution structure with convolution steps of 1 and 2, and finally, each layer has 128-dimensional features. For the cross sequence, after the sequence is spliced with the cross-time, the convolution pooling layer with the same structure as the link is used to perform convolution and pooling once, with a step size of 1. After convolution and pooling, 384-dimensional features are obtained, and the feature interacts through $MLP$\cite{muerle1963project}.

In the deep structure, we pass the slice-id, driver-id and other category features through the embedding layer to interact with the dense features using $MLP$. Finally, the convolutional features, the interactive features, link-time, link-ratio, cross-time, the original distance and the simple-eta are spliced together, and passed through a layer of $MLP$, moreover the $MAPE$ loss is used for loss calculation and backpropagation. The network structure is indicated in Figure 3.
\subsubsection{Tree model scheme}
The tree model uses the commonly used boosting integrated algorithm Gradient Boosting Tree (GBDT)\cite{friedman2001greedy}. The base learner is integrated through serial, and the performance of the boosting model can often be better than the base learner. Given a data set $D = \left \{ x_{i},y_{i} \right \}$, GBDT learns $K$ trees serially, and uses the following function to make predictions:

\begin{equation}
    \widehat{y}=\Phi (x_{i})=\sum_{k=1}^{K} f_{k}(x_{i}),f_{k}\in F
\end{equation}
 $f \left ( x \right )$ is a regression tree, and  $F$ is a collection of trees. The following tree fits the pseudo-residuals of all the previous trees and continuously approximates the real value through integration. The objective function in the training process is:
 
 \begin{equation}
     L\left ( \Phi  \right )= l\left (  \Phi \right ) + \Omega \left (  \Phi\right ) = \sum_{i} l\left ( y_{i},\widehat{y_{i}} \right ) + \sum_{k} \Omega \left (  f_{k}\right )
 \end{equation}

Among them, $l \left ( \Phi \right )$ is the error term and $l$ is the loss function. Commonly used losses for regression problems are $MSE$ and $MAE$. $\Omega \left ( \Phi \right )$ is the regular term, and the regular term penalizes the complexity of each regression tree.

There are already better and more commonly used second-order variants XGBOOST\cite{chen2016xgboost}, Lightgbm\cite{ke2017lightgbm}, and Catboost\cite{prokhorenkova2017catboost}. For this question, we used Lightgbm and Catboost for modeling and MAE as the loss function. Through feature engineering, 371 tree model features were finally constructed, and the prediction results of Lightgbm and Catboost were finally combined by weighted fusion.

\subsubsection{Ensemble}
There are many ways to ensemble. Here we use a simple weighted average method. First of all, for the neural network, we have trained a total of 8 models by modifying the size of the embedding and the truncation method of the link sequence. Since the link sequence of each order is not in the same length, we only keep the length of 200, so orders with a sequence greater than 200 will be truncated. At this time, the front part and the back part can be truncated. Our method is to use two truncated data. We train different models, and finally combine them in a weighted ensemble way. We also modify the length of embedding to make it a different model. The embedding size is set to 9, 15, 20, 30, and each size is trained once for different truncation methods. In the end, a total of 8 neural networks are trained, and the forward truncated embedding with an embedding size of 9 is input to the tree model. The final ensemble is shown in Figure 4.

\section{EXPERIMENT}
We used the method described above for modeling. Finally, we got a score of 0.11981 on the A list, and a score of 0.11974 on the B list, ranking first. Furthermore, the difference between the scores of A and B is relatively small, indicating that our program has a good generalization ability.
\section{CONCLUSION}
In this paper, we propose a novel solution to the ETA estimation problem, which is to combine the tree model with the neural network of the CNN structure. This scheme has strong accuracy and robustness on both A and B lists. In the end, we won the first place in the SIGSPATIAL 2021 GISCUP competition.

\bibliographystyle{ACM_Reference_Format}
\bibliography{sample_sigconf.bib}

\appendix

\end{document}